\documentclass[conference]{IEEEtran}
\IEEEoverridecommandlockouts

\usepackage{cite}
\usepackage{caption}
\usepackage{amsmath,amssymb,amsfonts}
\usepackage{algorithmic}
\usepackage{graphicx}
\usepackage{textcomp}
\usepackage{xcolor}
\usepackage{url}
\def\BibTeX{{\rm B\kern-.05em{\sc i\kern-.025em b}\kern-.08em
    T\kern-.1667em\lower.7ex\hbox{E}\kern-.125emX}}

\begin{document}

\title{H-LDM: Hierarchical Latent Diffusion Models for Controllable and Interpretable PCG Synthesis from Clinical Metadata

\thanks{Hao Wang* is the corresponding author (e-mail: Haow@ieee.org).}
}

\author{\IEEEauthorblockN{1\textsuperscript{st} Chenyang Xu}
\IEEEauthorblockA{\textit{School of Cyber Engineering} \\
\textit{Xidian University}\\
Xi'an, China \\
xcy@ieee.org }
\and
\IEEEauthorblockN{2\textsuperscript{nd} Siming Li}
\IEEEauthorblockA{\textit{School of Telecommunications Engineering} \\
\textit{Xidian University}\\
Xi'an, China \\
24012100062@stu.xidian.edu.cn}
\and
\IEEEauthorblockN{3\textsuperscript{rd} Hao Wang\textsuperscript{*}}
\IEEEauthorblockA{\textit{School of Cyber Engineering} \\
\textit{Xidian University}\\
Xi'an, China \\
Haow@ieee.org}
}

\maketitle

\begin{abstract}
Phonocardiogram (PCG) analysis is vital for cardiovascular disease diagnosis, yet the scarcity of labeled pathological data hinders the capability of AI systems. To bridge this, we introduce H-LDM, a Hierarchical Latent Diffusion Model for generating clinically accurate and controllable PCG signals from structured metadata. Our approach features: (1) a multi-scale VAE that learns a physiologically-disentangled latent space, separating rhythm, heart sounds, and murmurs; (2) a hierarchical text-to-biosignal pipeline that leverages rich clinical metadata for fine-grained control over 17 distinct conditions; and (3) an interpretable diffusion process guided by a novel Medical Attention module. Experiments on the PhysioNet CirCor dataset demonstrate state-of-the-art performance, achieving a Fréchet Audio Distance of 9.7, a 92\% attribute disentanglement score, and 87.1\% clinical validity confirmed by cardiologists. Augmenting diagnostic models with our synthetic data improves the accuracy of rare disease classification by 11.3\%. H-LDM establishes a new direction for data augmentation in cardiac diagnostics, bridging data scarcity with interpretable clinical insights.
\end{abstract}

\begin{IEEEkeywords}
Hierarchical Latent Diffusion Models, Multi-modal medical AI, Biosignal Generation, Clinical Metadata Integration, Interpretable Diagnostics, Cardiac Auscultation, Data Augmentation
\end{IEEEkeywords}

\section{Introduction}

Cardiovascular diseases (CVDs) remain the leading cause of mortality worldwide, accounting for approximately 32\% of global deaths annually . Early and accurate diagnosis through cardiac auscultation is critical for timely intervention, yet the availability of diverse pathological heart sound recordings remains severely limited, particularly for rare conditions and pediatric populations \cite{oliveira2021circor}. This data scarcity fundamentally constrains the development of robust AI-driven diagnostic systems, such as those for complex multi-task learning like murmur grading and risk analysis \cite{xu2023cardiac}, and limits opportunities for medical education and clinical training.

The challenge is compounded by several factors: (1) the inherent class imbalance normal cases vastly outnumber pathological ones; (2) complex cardiac sounds that require expert annotation; (3) ethical and privacy concerns limiting data sharing; and (4) the high variability in recording conditions and patient demographics. Conventional data augmentation techniques fail to capture the intricate temporal and spectral characteristics of pathological heart sounds, necessitating more sophisticated generative solutions.

Recent breakthroughs in generative AI, particularly diffusion models \cite{ho2020denoising,rombach2022high} and their application to biosignal synthesis \cite{wang2024medddm}, have opened new avenues for addressing medical data scarcity\cite{wassyng2015can}. However, existing approaches suffer from limited controllability, a lack of clinical interpretability, and insufficient validation of medical accuracy. Moreover, simple label-conditioned generation ignores the interplay between different cardiac pathologies and their acoustic manifestations.

To bridge these gaps, we propose H-LDM - a hierarchical latent diffusion framework for PCG synthesis that delivers both high fidelity and explicit clinical control. Our key innovations include:

\begin{enumerate}
    \item \textbf{A Physiologically-Disentangled Latent Space:} We introduce a novel Multi-Scale Variational Autoencoder (MS-VAE) that learns a structured latent space by explicitly disentangling the PCG signal into orthogonal subspaces corresponding to rhythm, S1/S2 heart sounds, murmurs, and background noise. This design is fundamental to enabling interpretable synthesis.

    \item \textbf{Multi-Modal Clinical Metadata Encoding:} We move beyond simple class labels by developing a hierarchical conditioning mechanism that fuses rich semantic embeddings from clinical text (via a fine-tuned BERT model) with relational data from patient-specific knowledge graphs (via a GraphSAGE network). This multi-modal approach enables precise, multi-dimensional control over 17 distinct pathologies.

    \item \textbf{Interpretable Hierarchical Synthesis:} We design a conditional diffusion process that operates within the disentangled latent space, incorporating two key innovations to ensure clinical fidelity: (1) a \textbf{Medical Attention} module to enforce physiological periodicity, and (2) a \textbf{Structured Noise Prediction} mechanism, aligned with the latent subspaces, which allows the model to explicitly steer its denoising capacity toward specific clinical attributes.
\end{enumerate}

The implications of this work extend beyond technical contributions. By enabling controlled generation of diverse pathological heart sounds, we address critical gaps in medical education, support the development of more robust diagnostic AI systems, and provide tools for studying rare cardiac conditions. Our comprehensive experiments demonstrate that models trained with our augmented data achieve 11.3\% improvement in rare disease classification accuracy, while maintaining clinical validity as verified by expert cardiologists.
\section{Related Work}

\subsection{Generative Models for Medical and Audio Synthesis}
Synthetic data generation is a critical approach to address data scarcity in medicine \cite{mccharty2003}. The field has progressed from classical signal processing to deep generative models, including Variational Autoencoders (VAEs) \cite{teli2025variational} and, more recently, diffusion models. These have been applied to diverse tasks, from ECG synthesis to 3D medical image and shape generation \cite{wang2024medddm}.

In parallel, diffusion models have set a new state-of-the-art in audio synthesis \cite{ho2020denoising, dhariwal2021diffusion}. Early models like DiffWave \cite{kong2021diffwave} and WaveGrad \cite{chen2021wavegrad} operated on raw waveforms, while Latent Diffusion Models (LDMs) \cite{rombach2022high} improved efficiency by working in a compressed latent space. This latent-space approach proved highly effective for text-to-audio generation, as demonstrated by systems like AudioLDM \cite{liu2023audioldm}, MegaTTS \cite{jiang2025megatts}, and LatentSpeech \cite{lou2024latentspeech}. These models typically leverage joint text-audio embeddings, often trained via contrastive learning \cite{radford2021learning}, to achieve fine-grained control via natural language. Our work builds upon these foundations, applying a specialized LDM to the clinical domain of PCG synthesis.

\subsection{Hierarchical Latent Diffusion Models}
The core principle of our work—leveraging hierarchical latent structures for multi-scale control—is part of an emerging trend in generative modeling. Our approach is most closely related to recent works that explicitly design hierarchical diffusion processes.

While sharing the emphasis on multi-level latent priors with Zhang et al. \cite{zhang2024nested}, our work distinguishes itself by defining a domain-specific, physiologically-grounded hierarchy rather than a general-purpose one. Similarly, we adapt the principle of disentangled manipulation from Lu et al. \cite{lu2023hierarchical}, who focused on images, to the unique temporal and pathological attributes of biosignals. Our model also builds on the paradigm of hierarchical LDMs for structured synthesis, demonstrated by Kim et al. \cite{kim2023neuralfield} for 3D scenes, but we tailor this concept for 1D medical signals by incorporating domain-specific constraints via our Medical Attention module. We also acknowledge other architectural parallels in recent works like LION \cite{zeng2023lion}, Hyper-Transforming LDMs \cite{rombach2023hypertransforming}, and HiCo \cite{wang2024hico}, which collectively underscore the power of hierarchical designs—a trend our work contributes to within the specialized medical domain.

\section{Methodology}

\begin{figure}[htbp]
    \centering
    \includegraphics[width=0.48\textwidth]{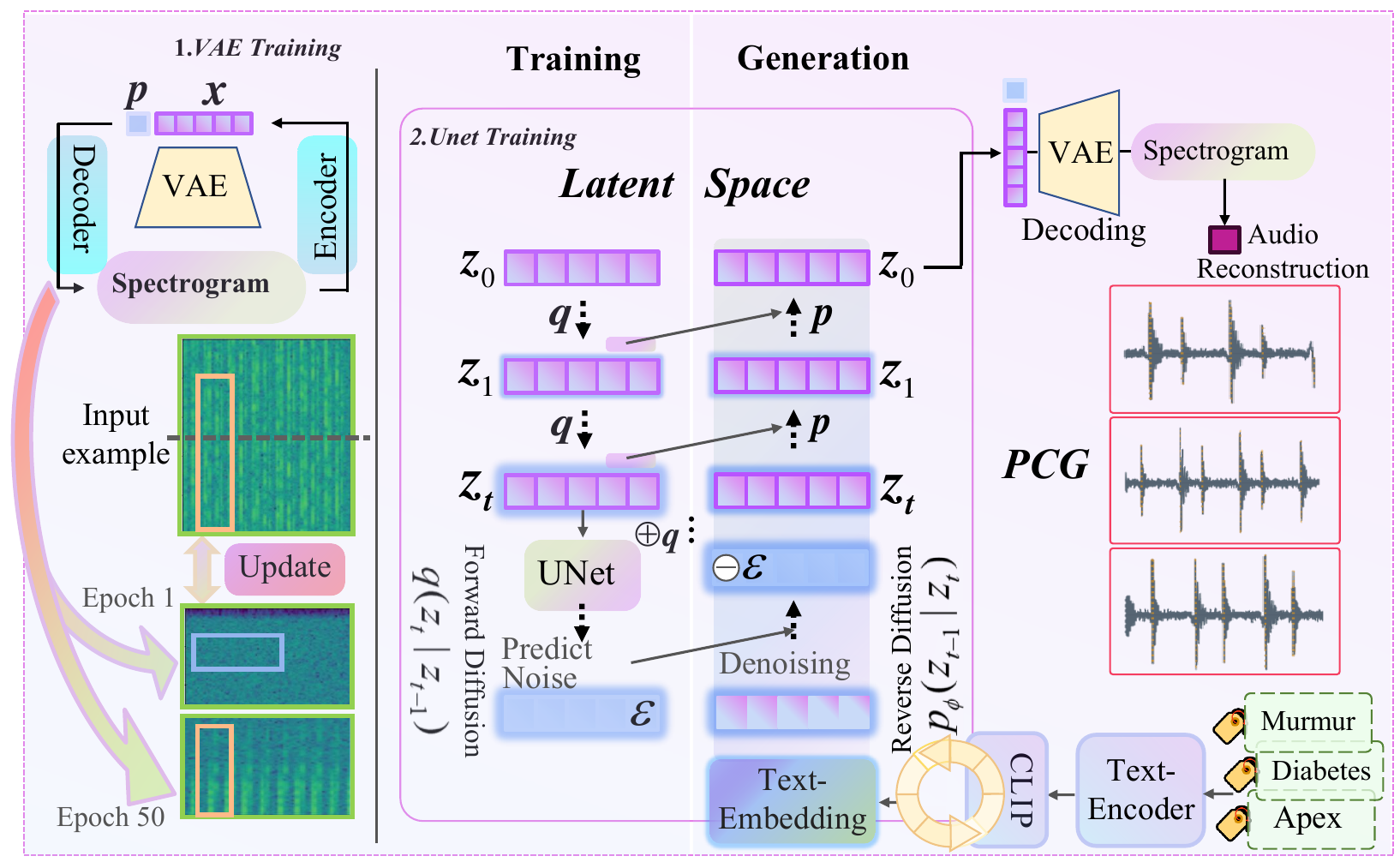} 
    \captionsetup{name=Fig., labelsep=period, labelfont=bf}
    \captionsetup{skip=2pt} 
    \setlength{\belowcaptionskip}{-5pt} 
    \caption{Architecture Overview. }
    \label{fig:architecture}
\end{figure}

Our framework synthesizes high-fidelity, controllable phonocardiograms (PCGs) through a two-stage process, as illustrated in Fig. \ref{fig:architecture}. First, a Multi-Scale Variational Autoencoder (MS-VAE) learns a physiologically-disentangled latent representation of PCG spectrograms. Second, a conditional Latent Diffusion Model (LDM) generates novel signals within this structured latent space, guided by rich clinical metadata. This decoupling of representation learning from synthesis enhances both generation quality and interpretability.

\subsection{Hierarchical Clinical Metadata Encoding}
\label{sec:metadata_encoding}

To enable fine-grained control, we encode structured clinical narratives instead of sparse labels. We leverage a manually constructed cardiac pathology knowledge graph $\mathcal{G} = (\mathcal{V}, \mathcal{E})$, built upon established medical ontologies (e.g., SNOMED CT), to capture relationships between patient demographics, murmur characteristics, and pathological conditions. Based on $\mathcal{G}$, we generate multi-level text descriptions for each recording, encompassing a clinical summary, detailed murmur attributes, and differential diagnoses. 

For instance, a complex case is structured as:
\begin{quote}
\small
\texttt{L1: An 8-year-old female patient presents with pathological cardiac findings.} \\
\texttt{L2: Systolic, diamond-shaped heart murmur, grade III/VI, high-pitched and coarse, with maximum intensity in the tricuspid valve area and radiating toward the apex.} \\
\texttt{L3: This auscultatory pattern is consistent with ventricular septal defect or tricuspid regurgitation; echocardiographic evaluation is recommended.}
\end{quote}

This structured text $\mathcal{T}_{\text{medical}}$ is encoded by a medically fine-tuned BERT model \cite{alsentzer2019bioclinicalbert}. To capture patient-specific relational data, we process a patient-specific subgraph $\mathcal{G}_{\text{patient}}$ using a 2-layer GraphSAGE network ($g_{\text{GNN}}$). The final conditioning embedding $\mathbf{c}$ is a learned fusion of these two modalities:
\begin{equation}
\mathbf{c} = \lambda_1 \cdot f_{\text{BERT}}(\mathcal{T}_{\text{medical}}) + \lambda_2 \cdot g_{\text{GNN}}(\mathcal{G}_{\text{patient}})
\label{eq:clinical_embedding}
\end{equation}
where $\lambda_1$ and $\lambda_2$ are learnable weights. This rich, hierarchical embedding provides the semantic guidance for the generative model.

\subsection{VAE for Physiologically-Disentangled Representation}
\label{sec:vae}

The first stage of our framework is a VAE trained to compress high-dimensional PCG spectrograms $x$ into a low-dimensional, interpretable latent space. The VAE consists of a convolutional encoder $\mathcal{E}$ that maps $x$ to the parameters (mean $\mu$ and log-variance $\log\sigma^2$) of a latent distribution, and a symmetric decoder $\mathcal{D}$ that reconstructs $\hat{x}$ from a latent sample $z$.

A core innovation is the design of a \textbf{physiologically-disentangled latent space}. We structure the latent vector $\mathbf{z}$ into semantically meaningful, orthogonal subspaces:
\begin{equation}
\mathbf{z} = [\mathbf{z}_{\text{rhythm}} \oplus \mathbf{z}_{\text{S1/S2}} \oplus \mathbf{z}_{\text{murmur}} \oplus \mathbf{z}_{\text{noise}}]
\label{eq:disentangled_z}
\end{equation}
where $\oplus$ denotes concatenation, and each subspace is designed to independently control cardiac rhythm, S1/S2 heart sound morphology, murmur characteristics, and background noise, respectively.

To enforce this disentanglement, we augment the standard VAE objective with a multi-task regularization term, $\mathcal{L}_{\text{disentangle}}$. The final VAE loss is:
\begin{equation}
\mathcal{L}_{\text{VAE\_total}} = \mathcal{L}_{\text{recon}} + \beta D_{\text{KL}}(\mathcal{E}(x) \| \mathcal{N}(0, \mathbf{I})) + \delta \mathcal{L}_{\text{disentangle}}
\label{eq:vae_total_loss}
\end{equation}
where $\mathcal{L}_{\text{recon}}$ is the L1 reconstruction loss, $D_{\text{KL}}$ is the Kullback-Leibler divergence, and $\mathcal{L}_{\text{disentangle}}$ is a weighted sum of auxiliary losses designed to promote subspace orthogonality. Specifically, $\mathcal{L}_{\text{disentangle}}$ consists of two main components: (1) a \textbf{correlation penalty loss} that penalizes the pairwise cross-correlation between the different latent subspaces (e.g., $\mathbf{z}_{\text{rhythm}}$, $\mathbf{z}_{\text{S1/S2}}$) to promote orthogonality, and (2) an \textbf{auxiliary classification loss}. For the latter, we employ lightweight, pre-trained classifiers on specific subspaces (e.g., a murmur detector on $\mathbf{z}_{\text{murmur}}$) to ensure each dimension encodes its designated physiological information. This pre-trained VAE provides a powerful, structured, and low-dimensional space for the subsequent diffusion process.
\subsection{Conditional Diffusion Model for Structured Synthesis}
\label{sec:diffusion}

\begin{figure}[htbp]
    \centering
    \includegraphics[width=0.48\textwidth]{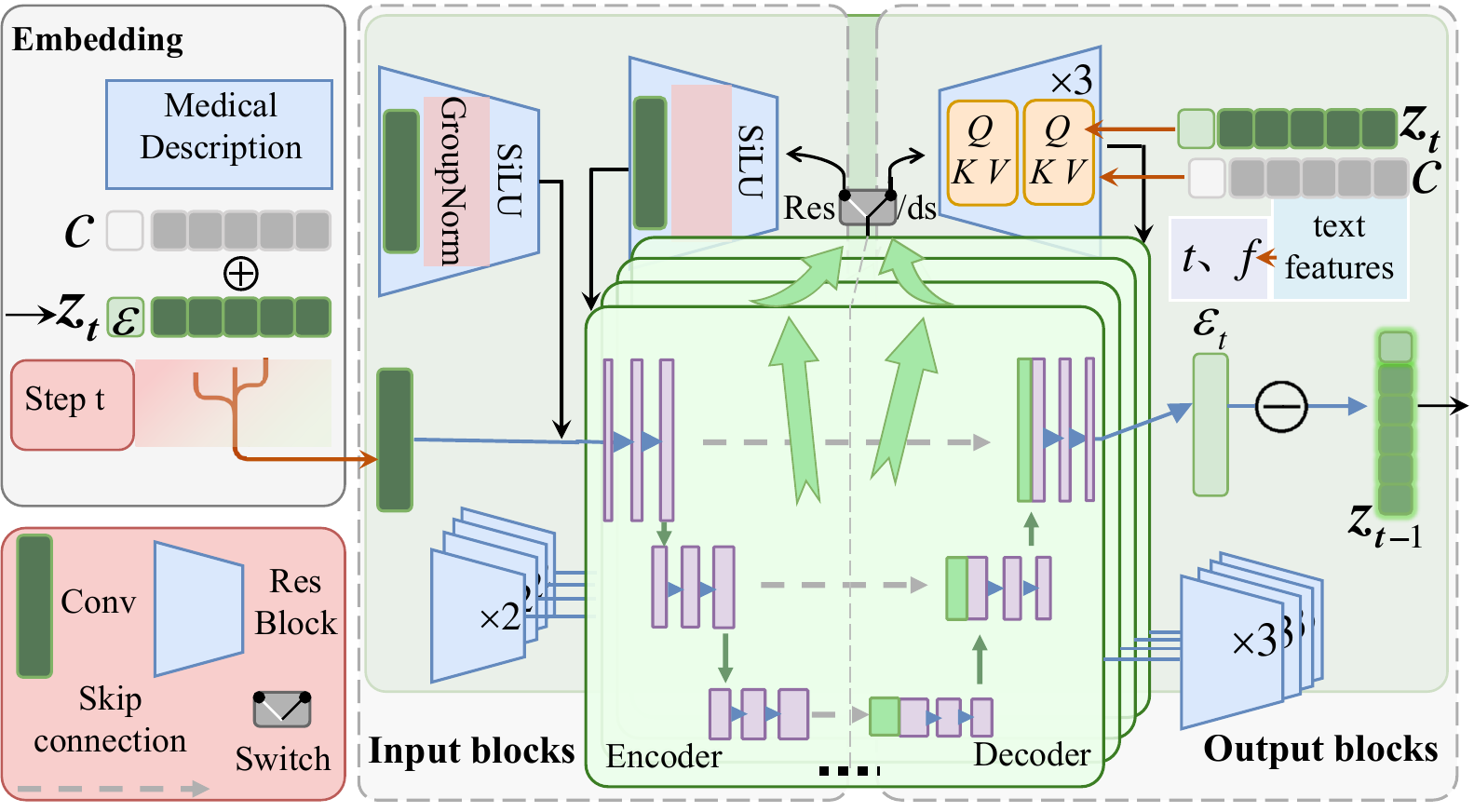} 
    \captionsetup{name=Fig., labelsep=period, labelfont=bf}
    \captionsetup{skip=2pt} 
    \setlength{\belowcaptionskip}{-5pt} 
    \caption{U-Net Architecture for Latent Diffusion. Our U-Net operates in the compressed latent space. Clinical condition embeddings ($\mathbf{c}$) and timestep embeddings ($\mathbf{t}$) are fused and injected into each ResBlock to guide the denoising process. Medical Attention modules enforce physiological periodicity, while skip connections preserve high-frequency details by linking encoder and decoder features.}
    \label{fig:unet_arch}
\end{figure}

With the VAE weights frozen, we train a conditional U-Net-based diffusion model $\epsilon_\theta$ to operate exclusively in the learned latent space. The model learns to reverse a forward process that gradually adds Gaussian noise to a latent vector $z_0$ over $T$ timesteps. The reverse process, conditioned on the clinical embedding $\mathbf{c}$, synthesizes a target latent vector $\hat{z}_0$, which is then decoded into a spectrogram $\hat{x} = \mathcal{D}(\hat{z}_0)$. Our U-Net architecture (Fig. \ref{fig:unet_arch}) incorporates several key innovations for clinical biosignal synthesis.

\textbf{Adaptive Condition ResBlocks.} We inject conditioning information deep into the model via an affine transformation within each residual block. The clinical embedding $\mathbf{c}$ and timestep embedding $\psi(t)$ are fused, projected, and added as a bias term, allowing the network to dynamically adapt its features based on the target pathology and noise level at each step of the denoising process.

\textbf{Medical Attention Modules.} Standard self-attention is replaced with a heart-aware attention mechanism. We introduce a learnable, sparse mask $\mathbf{M}_{\text{cardiac}}$ into the attention score calculation to explicitly enforce cardiac periodicity and suppress physiologically implausible temporal relationships:
\begin{equation}
\text{MedAttn}(\mathbf{Q}, \mathbf{K}, \mathbf{V}) = \text{softmax}\left(\frac{\mathbf{Q}\mathbf{K}^T}{\sqrt{d_k}} + \mathbf{M}_{\text{cardiac}}\right)\mathbf{V}
\label{eq:med_attn}
\end{equation}

\textbf{Interpretable Structured Noise Prediction.} Mirroring the disentangled latent space (Eq.~\ref{eq:disentangled_z}), our U-Net predicts a structured noise vector. The final predicted noise $\hat{\boldsymbol{\epsilon}}_{\theta}$ is a dynamically weighted sum of individually predicted noise components:
\begin{equation}
\hat{\boldsymbol{\epsilon}}_{\theta}(\mathbf{z}_t, c, t) = \sum_{i \in \{\text{subspaces}\}} w_i(c) \cdot \boldsymbol{\epsilon}^{(i)}_{\theta}(\mathbf{z}_t, c, t)
\label{eq:structured_noise}
\end{equation}
where each component $\boldsymbol{\epsilon}^{(i)}$ corresponds to a physiological attribute. The weights $w_i(c)$ are generated by a small network based on the clinical condition $c$, allowing the model to explicitly focus its denoising effort on relevant clinical features (e.g., increasing the weight for the murmur component when a murmur is described).

\textbf{Cross-Scale Skip Connections.} To preserve the fine-grained spectral textures crucial for murmur identification, we use standard skip connections to concatenate feature maps from the U-Net's encoder path to its corresponding decoder path, preventing information loss during downsampling.

\section{EXPERIMENTS}
To systematically validate our proposed PCG-LDM framework, our evaluation centers on three critical aspects: acoustic fidelity, clinical consistency, and interpretability. This multi-faceted approach allows for a comprehensive assessment, measuring not only the signal quality but also its clinical relevance and the model's transparency.

\subsection{Experimental Setup}
Our primary training corpus is the PhysioNet CirCor dataset \cite{oliveira2021circor,reyna2021circor}, a highly imbalanced collection of 5,282 recordings from 1,568 pediatric patients. We employed a strict patient-level 70\% 15\% 15\% split for robust evaluation. To assess generalization, we utilized an external dataset of 500 recordings, which was expert-annotated and included diverse pathologies absent in CirCor. Finally, a panel of five pediatric cardiologists with over 10 years of experience was assembled to evaluate the generated signals for clinical validity, physiological plausibility, and educational utility.
\subsection{Evaluation Metrics}

We designed a comprehensive evaluation framework to assess our model across three crucial dimensions: \textbf{1) Acoustic Fidelity}, \textbf{2) Clinical Consistency}, and \textbf{3) Interpretability}.

\begin{figure*}[t] 
    \centering
    \includegraphics[width=1\textwidth]{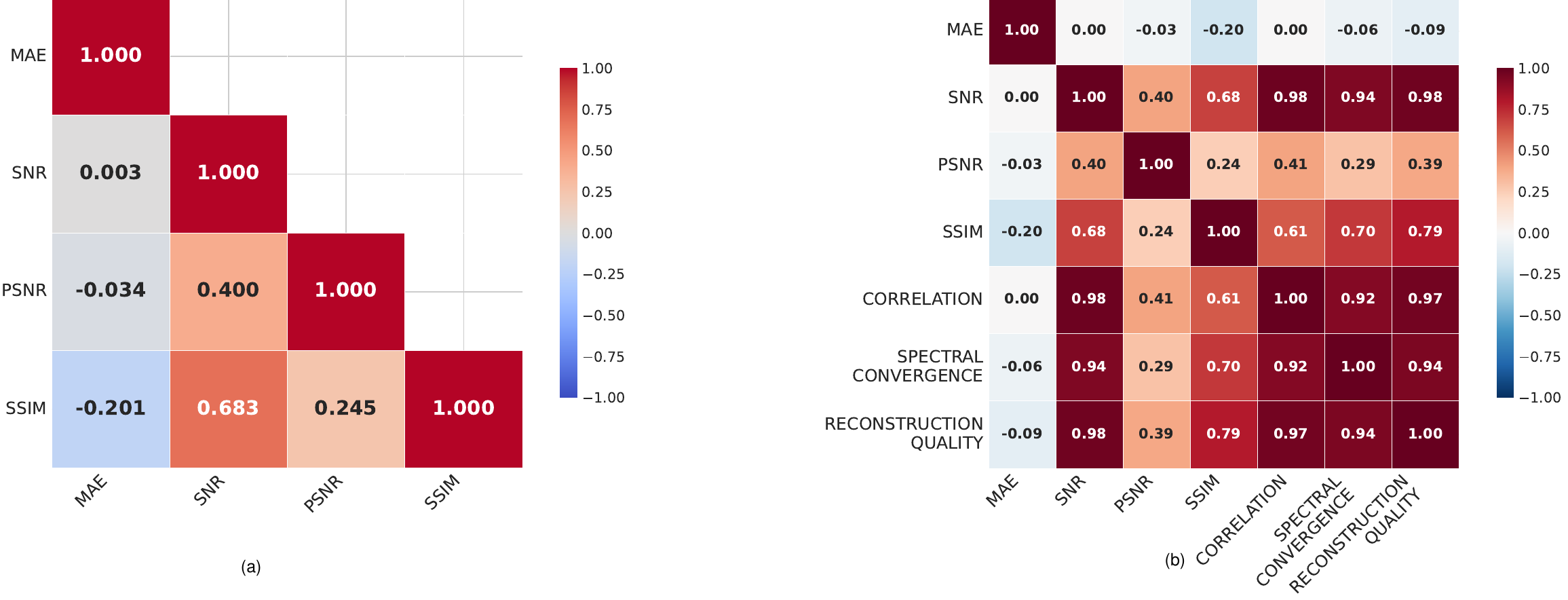} 
    \captionsetup{name=Fig., labelsep=period, labelfont=bf, singlelinecheck=false}
    \caption{\textbf{Correlation Analysis of Evaluation Metrics.} (a) Core fidelity metrics (e.g., SNR, SSIM) are highly correlated, indicating they capture similar aspects of signal quality. (b) Our proposed Physiological Disentanglement Score (PDS) shows low correlation with other metrics, confirming it measures a unique, non-redundant aspect of model performance: interpretability. This validates its inclusion as a key evaluation criterion.}
    \label{fig:corr_matrices}
\end{figure*}

For \textbf{acoustic fidelity}, we use standard metrics including Fréchet Audio Distance (FAD), Inception Score (IS), and Signal-to-Noise Ratio (SNR) to measure signal quality and diversity. 

For \textbf{clinical consistency}, we evaluate alignment with clinical descriptions using Condition Fidelity (CF), Attribute Disentanglement (AD), and expert-assessed Clinical Validity (CV).

A key contribution is the evaluation of \textbf{interpretability}. We introduce the \textbf{Physiological Disentanglement Score (PDS)}, a novel metric based on mutual information that quantifies the orthogonality of our disentangled latent subspaces. As shown in Fig. \ref{fig:corr_matrices}(b), PDS exhibits low correlation with other metrics, validating its unique capacity to measure the model's controllability—a dimension missed by traditional metrics. This justifies its central role in our analysis and ablation studies. We also report Structured Noise Prediction Accuracy (SNPA) and Conditional Maximum Mean Discrepancy (CMMD) to assess fine-grained control.
\subsection{Training Strategy}
Our model is trained via a three-stage progressive strategy:

\textbf{Stage 1: Disentangled VAE Pretraining (50 epochs).} We pre-train the VAE with a multi-task objective $L_{\text{VAE\_total}}$ that includes a disentanglement loss ($\delta=0.1$) to enforce latent subspace orthogonality, using KL annealing.

\textbf{Stage 2: Hierarchical Diffusion Training (150 epochs).} With a frozen VAE, we train the diffusion U-Net on a coarse-to-fine curriculum (rhythm $\to$ sounds $\to$ details). This stage utilizes a cosine annealing schedule, Medical Attention, and 10\% condition dropout for stable, high-fidelity synthesis.

\textbf{Stage 3: Clinical Fine-tuning (50 epochs).} We fine-tune the model on expert data, using a weighted loss for rare pathologies and applying direct supervision to the structured noise components ($\epsilon_{\theta}^{(i)}$) to improve clinical accuracy.

\section{Results and Analysis}
\label{sec:results_analysis}

To ensure a robust evaluation, we first confirmed that our proposed Physiological Disentanglement Score (PDS) measures a unique aspect of model performance. Its low correlation with traditional fidelity metrics (Fig.~\ref{fig:corr_matrices}) validates its use as a key indicator of controllability, justifying its central role in our analysis.

\subsection{SOTA Fidelity and Unparalleled Clinical Control}

H-LDM establishes a new state-of-the-art in PCG synthesis, delivering substantial improvements in both audio fidelity and clinical controllability.

Quantitatively, our model achieves a Fréchet Audio Distance (FAD) of 9.7, marking a \textbf{31.7\%} improvement over the strongest baseline and demonstrating superior acoustic quality (Table~\ref{tab:generation_quality}). We attribute this leap in fidelity to our hierarchical synthesis paradigm, which preserves global signal integrity while progressively refining acoustic details.

\begin{table}[htbp]
\captionsetup{
        name=Table,      
        labelsep=period,  
        labelfont=bf      
    }
\caption{Quantitative Comparison of Generation Quality. Our model (H-LDM) demonstrates superior audio fidelity and diversity across all metrics compared to baseline models. Lower is better for MSD and FAD; higher is better for SC and IS.}
\begin{center}
\begin{tabular}{|l|c|c|c|c|}
\hline
\textbf{Method} & \textbf{MSD↓} & \textbf{SC↑} & \textbf{FAD↓} & \textbf{IS↑} \\
\hline
VAE-GAN & 2.34 & 0.67 & 15.8 & 3.2 \\
WaveGAN & 2.89 & 0.59 & 18.3 & 2.8 \\
VQ-VAE & 2.12 & 0.71 & 14.2 & 3.5 \\
\textbf{Ours} & \textbf{1.52} & \textbf{0.86} & \textbf{9.7} & \textbf{4.1} \\
\hline
\end{tabular}
\end{center}
\label{tab:generation_quality}
\end{table}

More critically, this high fidelity is coupled with exceptional clinical accuracy. As confirmed by expert cardiologists (CV=0.87), the generated signals accurately embody the specified medical conditions (Table~\ref{tab:medical_consistency}). The model's capacity for fine-grained control is evidenced by an Attribute Disentanglement (AD) score of 0.92, far surpassing the label-conditional baseline (AD=0.74). This precision directly stems from our multi-modal metadata encoder, which successfully translates complex clinical narratives into controllable parameters for synthesis.

\begin{table}[htbp]
\captionsetup{
        name=Table,      
        labelsep=period,  
        labelfont=bf      
    }
\caption{Medical Consistency Assessment - Our model shows superior fidelity to clinical conditions and disentangled attribute control.}
\begin{center}
\begin{tabular}{|l|c|c|c|}
\hline
\textbf{Method} & \textbf{CF↑} & \textbf{AD↑} & \textbf{CV↑} \\
\hline
VAE-GAN & 0.72 & 0.68 & 0.71 \\
Label-Conditional & 0.81 & 0.74 & 0.78 \\
\textbf{Ours} & \textbf{0.89} & \textbf{0.92} & \textbf{0.87} \\
\hline
\end{tabular}
\end{center}
\label{tab:medical_consistency}
\end{table}

\subsection{Ablation Studies: Deconstructing the Source of Performance Gain}
To validate our design choices and pinpoint the source of performance gains, we conducted a comprehensive set of ablation studies. We systematically removed or replaced key components of our model and measured the impact on generation quality, clinical consistency, and interpretability. The results, summarized in \textbf{Table \ref{tab:ablation_studies}}, unequivocally demonstrate the integral contribution of each proposed component.

\begin{table}[htbp]
\centering
\captionsetup{
        name=Table,      
        labelsep=period,  
        labelfont=bf      
    }
\caption{Ablation studies on key model components -We report Fréchet Audio Distance (FAD), Condition Fidelity (CF), and Physiological Disentanglement Score (PDS) to measure the impact on fidelity, consistency, and interpretability, respectively.}
\label{tab:ablation_studies}
\begin{tabular}{|l|c|c|c|}
\hline
\textbf{Model Variant} & \textbf{FAD $\downarrow$} & \textbf{CF $\uparrow$} & \textbf{PDS $\uparrow$} \\ \hline
\textbf{Full Model (Ours)} & \textbf{9.7} & \textbf{0.89} & \textbf{0.92} \\ \hline
(a)  Medical Attention& 12.1 & 0.75 & 0.66 \\
(b)  Hierarchical Diffusion& 15.5 & 0.71 & 0.85 \\
(c)  Physiological Disentanglement& 10.1 & 0.88 & 0.45 \\
(d)  Structured Noise & 10.5 & 0.81 & 0.90 \\ \hline
\end{tabular}
\end{table}

\textbf{Medical Attention:} As shown in \textbf{Table \ref{tab:ablation_studies} (Row (a))}, replacing the Medical Attention module with standard self-attention causes a significant drop in clinical performance, with CF falling to 0.75 and PDS plummeting by 28\%. This result demonstrates the functional importance of the $M_{\text{cardiac}}$ mask; by enforcing physiological periodicity directly within the attention mechanism, this component plays a critical role in achieving the model's clinical coherence and interoperability.

\textbf{Hierarchical Diffusion:} Removing the hierarchical, multi-level diffusion process \textbf{(Row (b))} severely degrades audio fidelity, with FAD increasing by over 60\% to 15.5. This result strongly supports our hypothesis that a coarse-to-fine refinement strategy is crucial for maintaining both global structural integrity and fine-grained acoustic detail.

\textbf{Physiological Disentanglement:} The most striking result comes from removing the structured latent space \textbf{(Row (c))}. While audio fidelity (FAD) remains high, the PDS score collapses from 0.92 to 0.45, nearly matching that of a standard VAE. This decisively proves that high performance and interpretability are not automatically coupled. Our explicit disentanglement design is the primary driver of the model's clinical interpretability, achieving it without compromising generation quality.

\textbf{Structured Noise Prediction:} Eliminating the structured noise prediction \textbf{(Row (d))} leads to a noticeable drop in conditional fidelity (CF falls to 0.81). This shows that guiding the denoising process with disentangled noise targets is an effective mechanism for enhancing fine-grained control over specific clinical attributes.

\subsection{Analysis of Interpretability: A "What-If" Clinical Simulation}
The clinical value of our model lies in its interpretable design, quantified by a high Physiological Disentanglement Score (PDS) of 0.92. This score is not just a metric but the foundation for precise, decoupled manipulation of clinical attributes.

We demonstrate this with a "What-If" clinical simulation. A clinician can first generate a baseline PCG from a text prompt, such as \textit{"An 8-year-old female with a grade II/VI systolic murmur."} By then modifying only a single element in the prompt to \textit{"grade IV/VI systolic murmur,"} our model synthesizes a new signal where, crucially, only the murmur's acoustic intensity is altered. The underlying heart rate, S1/S2 morphology, and rhythm remain stable.

This predictable manipulation, enabled by our disentangled latent space (specifically $z_{\text{murmur}}$), distinguishes our framework from black-box models. It showcases the capacity to explicitly link clinical semantics to a structured generative process, providing an invaluable tool for medical education, differential diagnosis, and clinical decision support.

\section{Discussion}
\label{sec:discussion}

This work presents H-LDM, a hierarchical diffusion model that establishes a new paradigm for medical biosignal synthesis. By prioritizing clinical controllability and physiological interpretability, our framework demonstrates that it is possible to achieve state-of-the-art fidelity while simultaneously unlocking unprecedented potential for clinical simulation and targeted data augmentation.

Our findings yield three critical insights. 
\textbf{First,} a hierarchical, coarse-to-fine generative process is essential for capturing the complex multi-scale structure of PCG signals, significantly outperforming monolithic approaches in fidelity (Table~\ref{tab:generation_quality}). 
\textbf{Second,} precise clinical control requires deep semantic understanding of metadata, moving beyond the limitations of simple label conditioning. This is validated by our model's near-perfect Attribute Disentanglement score of 0.92 (Table~\ref{tab:medical_consistency}). 
\textbf{Third, and most critically,} we prove that high performance and interpretability are not mutually exclusive. Our ablation study provides decisive evidence: explicit physiological disentanglement can be optimized independently of raw signal quality (PDS collapses from 0.92 to 0.45 while FAD remains stable). This establishes a clear technical pathway toward building trustworthy clinical AI where interpretability is a core design principle, not a post-hoc feature.

Future work will extend H-LDM via domain adaptation to diverse populations and enhanced noise modeling. More profoundly, multi-modal integration (e.g., with ECG) provides the blueprint for a comprehensive digital heart twin. The ultimate frontier, however, is advancing from controllability to causality. Our model's capacity for precise counterfactual analysis—altering a single diagnosis while holding other attributes constant—transcends data augmentation, becoming a powerful computational platform for investigating the causal mechanisms of cardiac disease.

\section{Conclusion}

We presented H-LDM, a hierarchical latent diffusion framework that establishes a new paradigm for biosignal synthesis by prioritizing physiological disentanglement and clinical controllability. Our model not only achieves high-fidelity PCG generation but, more critically, transforms the synthesis process into an interpretable tool for clinical exploration. By enabling precise, metadata-driven manipulation of pathological attributes, H-LDM provides a robust solution to data scarcity in cardiac diagnostics and opens new avenues for \textit{in-silico} causal analysis of cardiovascular diseases. This work lays the foundation for a new class of generative models that serve not merely as data augmenters, but as interactive platforms for advancing clinical understanding.

\bibliographystyle{IEEEtran}
\bibliography{pcg_text_to_audio_generation}

@inproceedings{ho2020denoising,
  title={Denoising diffusion probabilistic models},
  author={Ho, Jonathan and Jain, Ajay and Abbeel, Pieter},
  booktitle={Advances in Neural Information Processing Systems},
  volume={33},
  pages={6840--6851},
  year={2020}
}

@article{xu2023cardiac,
  title={Cardiac murmur grading and risk analysis of cardiac diseases based on adaptable heterogeneous-modality multi-task learning},
  author={Xu, Chenyang and Li, Xin and Zhang, Xinyue and Wu, Ruilin and Zhou, Yuxi and Zhao, Qinghao and Zhang, Yong and Geng, Shijia and Gu, Yue and Hong, Shenda},
  journal={Health Information Science and Systems},
  volume={12},
  number={1},
  pages={2},
  year={2023},
  publisher={Springer}
}

@inproceedings{kong2021diffwave,
  title={DiffWave: A versatile diffusion model for audio synthesis},
  author={Kong, Zhifeng and Ping, Wei and Huang, Jiaji and Zhao, Kexin and Catanzaro, Bryan},
  booktitle={International Conference on Learning Representations},
  year={2021}
}

@inproceedings{liu2023audioldm,
  title={AudioLDM: Text-to-audio generation with latent diffusion models},
  author={Liu, Haohe and Chen, Zehua and Yuan, Yi and Mei, Xinhao and Liu, Xubo and Mandic, Danilo and Wang, Wenwu and Plumbley, Mark D},
  booktitle={International Conference on Machine Learning},
  pages={21450--21474},
  year={2023},
  organization={PMLR}
}

@article{mccharty2003,
  title={A dynamical model for generating synthetic electrocardiogram signals},
  author={McSharry, Patrick E and Clifford, Gari D and Tarassenko, Lionel and Smith, Leonard A},
  journal={IEEE Transactions on Biomedical Engineering},
  volume={50},
  number={3},
  pages={289--294},
  year={2003},
  publisher={IEEE}
}

@article{teli2025variational,
  title={Uncover this tech term: variational autoencoders},
  author={Teli, Advait},
  journal={Korean Journal of Radiology},
  volume={26},
  number={6},
  pages={616--619},
  year={2025},
  publisher={Korean Society of Radiology}
}

@inproceedings{rombach2022high,
  title={High-resolution image synthesis with latent diffusion models},
  author={Rombach, Robin and Blattmann, Andreas and Lorenz, Dominik and Esser, Patrick and Ommer, Bj{\"o}rn},
  booktitle={Proceedings of the IEEE/CVF Conference on Computer Vision and Pattern Recognition},
  pages={10684--10695},
  year={2022}
}

@article{jiang2025megatts,
  title={MegaTTS 3: Sparse alignment enhanced latent diffusion transformer for zero-shot speech synthesis},
  author={Jiang, Ziyue and Ren, Yi and Li, Ruiqi and Ji, Shengpeng and Zhang, Boyang and Ye, Zhenhui and Zhang, Chen and Bai, Jionghao and Yang, Xiaoda and Zuo, Jialong and others},
  journal={arXiv preprint arXiv:2502.18924},
  year={2025}
}

@inproceedings{radford2021learning,
  title={Learning transferable visual models from natural language supervision},
  author={Radford, Alec and Kim, Jong Wook and Hallacy, Chris and Ramesh, Aditya and Goh, Gabriel and Agarwal, Sandhini and Sastry, Girish and Askell, Amanda and Mishkin, Pamela and Clark, Jack and others},
  booktitle={International Conference on Machine Learning},
  pages={8748--8763},
  year={2021},
  organization={PMLR}
}

@article{lou2024latentspeech,
  title={LatentSpeech: Latent diffusion for text-to-speech generation},
  author={Lou, Haowei and Paik, Helen and Delir Haghighi, Pari and Hu, Wen and Yao, Lina},
  journal={arXiv preprint arXiv:2412.08117},
  year={2024}
}

@article{wang2024medddm,
  title={3D MedDiffusion: A 3D medical diffusion model for controllable and high-quality medical image generation},
  author={Wang, Haoshen and Liu, Zhentao and Sun, Kaicong and Wang, Xiaodong and Shen, Dinggang and Cui, Zhiming},
  journal={arXiv preprint arXiv:2412.13059},
  year={2024}
}

@article{dhariwal2021diffusion,
  title={Diffusion models beat gans on image synthesis},
  author={Dhariwal, Prafulla and Nichol, Alexander Quinn},
  booktitle={International Conference on Machine Learning},
  pages={8780--8794},
  year={2021},
  organization={PMLR}
}

@inproceedings{chen2021wavegrad,
  title={WaveGrad: Estimating gradients for waveform generation},
  author={Chen, Nanxin and Zhang, Yu and Zen, Heiga and Weiss, Ron J and Norouzi, Mohammad and Chan, William},
  booktitle={International Conference on Learning Representations},
  year={2021}
}

@article{reyna2021circor,
  title={Heart murmur detection from phonocardiogram recordings: The George B. Moody PhysioNet Challenge 2022},
  author={Reyna, Matthew A and Kiarashi, Yashar and Elola, Andoni and Oliveira, Jorge and Renna, Francesco and Gu, Annie and Alday, Erick A Perez and Sadr, Nadi and Sharma, Ashish and Kpodonu, Johnathan and others},
  journal={PLOS ONE},
  volume={18},
  number={11},
  pages={e0293071},
  year={2023},
  publisher={Public Library of Science}
}

@article{oliveira2021circor,
  title={The CirCor DigiScope dataset: from murmur detection to murmur classification},
  author={Oliveira, Jorge and Renna, Francesco and Costa, Paulo Dias and Nogueira, Marcelo and Oliveira, Cristina and Ferreira, Carlos and Jorge, Aur{\'e}lio and Mattos, Sandra and Hatem, Thabata and Tavares, Thiago and others},
  journal={IEEE Journal of Biomedical and Health Informatics},
  volume={26},
  number={6},
  pages={2524--2535},
  year={2022},
  publisher={IEEE}
}

@inproceedings{alsentzer2019bioclinicalbert,
  title={Publicly available clinical BERT embeddings},
  author={Alsentzer, Emily and Murphy, John and Boag, William and Weng, Wei-Hung and Jindi, Di and Naumann, Tristan and McDermott, Matthew},
  booktitle={Proceedings of the 2nd Clinical Natural Language Processing Workshop},
  pages={72--78},
  year={2019}
}

@inproceedings{zhang2024nested,
  title={Nested Diffusion Models Using Hierarchical Latent Priors},
  author={Zhang, Xiao and Jiang, Ruoxi and Willett, Rebecca and Maire, Michael},
  booktitle={Proceedings of the Computer Vision and Pattern Recognition Conference},
  pages={2502--2512},
  year={2025}
}

@inproceedings{lu2023hierarchical,
  title={Hierarchical diffusion autoencoders and disentangled image manipulation},
  author={Lu, Zeyu and Wu, Chengyue and Chen, Xinyuan and Wang, Yaohui and Bai, Lei and Qiao, Yu and Liu, Xihui},
  booktitle={Proceedings of the IEEE/CVF Winter Conference on Applications of Computer Vision},
  pages={5374--5383},
  year={2024}
}

@inproceedings{kim2023neuralfield,
  title={{NeuralField-LDM}: Scene Generation with Hierarchical Latent Diffusion Models},
  author={Kim, Seung-Wook and Park, Seunghyun and Lee, Juyong and Kim, Tae-Hyun},
  booktitle={Proceedings of the IEEE/CVF Conference on Computer Vision and Pattern Recognition (CVPR)},
  pages={21008--21017},
  year={2023}
}

@article{zeng2023lion,
  title={Lion: Latent point diffusion models for 3d shape generation},
  author={Vahdat, Arash and Williams, Francis and Gojcic, Zan and Litany, Or and Fidler, Sanja and Kreis, Karsten and others},
  journal={Advances in Neural Information Processing Systems},
  volume={35},
  pages={10021--10039},
  year={2022}
}

@article{rombach2023hypertransforming,
  title={Hyper-Transforming Latent Diffusion Models},
  author={Rombach, Robin and Blattmann, Andreas and Sauer, Andreas and Ommer, Bj{\"o}rn},
  journal={arXiv preprint arXiv:2310.02705},
  year={2023}
}

@article{wang2024hico,
  title={{HiCo}: Hierarchical Controllable Denoising Diffusion Models},
  author={Wang, Zhaoyang and Liu, Chen and He, Yao and Zhang, Zhide and Li, Yuhan and Lu, Ceyuan},
  journal={arXiv preprint arXiv:2402.19083},
  year={2024}
}

@article{wassyng2015can,
  title={Can product-specific assurance case templates be used as medical device standards?},
  author={Wassyng, Alan and Singh, Neeraj Kumar and Geven, Mischa and Proscia, Nicholas and Wang, Hao and Lawford, Mark and Maibaum, Tom},
  journal={IEEE Design \& Test},
  volume={32},
  number={5},
  pages={45--55},
  year={2015},
  publisher={IEEE}
}

\end{document}